\documentclass[pmlr]{jmlr}

\RequirePackage{graphicx}

\usepackage{dsfont}
\usepackage{algorithm}
\usepackage{algorithmic}
\usepackage{xcolor}
\usepackage{makecell}
\usepackage{framed}
\usepackage{mdframed}
\usepackage{multirow}
\graphicspath{ {images/} }
\usepackage{listings}
\usepackage{hhline}
\usepackage{soul}
\usepackage{booktabs}
\usepackage{caption}

\lstdefinestyle{mystyle}{
    backgroundcolor=\color{gray!15},   
    commentstyle=\color{codegreen},
    keywordstyle=\color{magenta},
    numberstyle=\tiny\color{codegray},
    stringstyle=\color{codepurple},
    basicstyle=\ttfamily\footnotesize,
    breakatwhitespace=false,         
    breaklines=true,                 
    captionpos=b,                    
    keepspaces=true,                 
    numbers=left,                    
    numbersep=5pt,                  
    showspaces=false,                
    showstringspaces=false,
    showtabs=false,                  
    tabsize=2
}
\usepackage{longtable}
\makeatletter
\def\set@curr@file#1{\def\@curr@file{#1}} 
\makeatother
\usepackage[load-configurations=version-1]{siunitx} 

\theorembodyfont{\upshape}
\theoremheaderfont{\scshape}
\theorempostheader{:}
\theoremsep{\newline}

\jmlrvolume{298}
\jmlryear{2025}
\jmlrworkshop{Machine Learning for Healthcare}

\title[QB-RAG]{The Geometry of Queries: Query-Based Innovations in
Retrieval-Augmented Generation for Healthcare QA}

\author{\Name{Eric Yang}\textsuperscript{ †}
        \Email{eryang@verily.com}\\
        \addr Verily Life Sciences\\
        Dallas, TX, USA 
        \AND
        \Name{Jonathan Amar}\textsuperscript{ †}
        \Email{jonathanamar@verily.com}\\
        \addr Verily Life Sciences\\
        Dallas, TX, USA 
        \AND
        \Name{Jong Ha Lee}
        \Email{jonghalee@verily.com}\\
        \addr Verily Life Sciences\\
        Dallas, TX, USA 
        \AND
        \Name{Bhawesh Kumar}
        \Email{bhawesh.hsph@gmail.com}\\
        \addr Verily Life Sciences\\
        Dallas, TX, USA 
        \AND
        \Name{Yugang Jia}
        \Email{yugang@verily.com}\\
        \addr Verily Life Sciences\\
        Dallas, TX, USA}

\begin{document}

\maketitle

\footnotetext[1]{† Equal contribution.}

\begin{abstract}
Deploying Large Language Models (LLMs) for healthcare question answering requires robust methods to ensure accuracy and reliability. This work introduces Query-Based Retrieval Augmented Generation (QB-RAG), a framework for enhancing Retrieval-Augmented Generation (RAG) systems in healthcare question-answering by pre-aligning user queries with a database of curated, answerable questions derived from healthcare content. A key component of QB-RAG is an LLM-based filtering mechanism that ensures that only relevant and answerable questions are included in the database, enabling reliable reference query generation at scale. We provide theoretical motivation for QB-RAG, conduct a comparative analysis of existing retrieval enhancement techniques, and introduce a generalizable, comprehensive evaluation framework that assesses both the retrieval effectiveness and the quality of the generated response based on faithfulness, relevance, and adherence to the guideline. Our empirical evaluation on a healthcare data set demonstrates the superior performance of QB-RAG compared to existing retrieval methods, highlighting its practical value in building trustworthy digital health applications for health question-answering.
\end{abstract}

\section{Introduction}

\label{sec:intro}

Large Language Models (LLMs) have demonstrated remarkable capabilities across a diverse set of natural language understanding and generation tasks \citep{openai2024gpt4, touvron2023llama, anil2023palm}. In healthcare, LLMs hold immense promise for developing conversational AI systems that can answer patient questions, offer personalized health advice, and improve access to care, particularly for underserved populations 
\citep{clusmann2023future, peng2023study, alowais2023revolutionizing, nori2023capabilities, singhal2023towards, tu2024conversational}.

However, applying LLMs in healthcare presents significant challenges in ensuring the accuracy, reliability, and adherence to the latest healthcare practices. The probabilistic nature of LLMs, combined with limitations and potential biases in their training data, can lead to  hallucinations. Additionally, the knowledge within an LLM is limited to the data it was trained on and may not reflect latest clinical practice. To address these challenges, further fine-tuning, instruction tuning \citep{wei2022finetuned} and additional reinforcement learning \citep{ouyang2022training} approaches can be considered. However, these post-training approaches still have limitations: datasets can be difficult and expensive to acquire, especially in healthcare, and the computation cost may be prohibitive.

Retrieval Augmented Generation (RAG) \citep{lewis2020retrieval} offers a promising alternative by grounding LLM responses on a curated knowledge base of vetted information. Unlike relying solely on the LLM's internal knowledge, RAG systems retrieve relevant information for a specific user query from this external knowledge base to inform the LLM's answer generation. This grounding helps mitigate hallucinations and allows developers to incorporate the latest knowledge. Applications using RAG systems have been shown to be quite compelling compared to traditional LLM fine-tuning \citep{gupta2024rag, ovadia2024finetuning}. However, the effectiveness of RAG depends critically on its ability to accurately retrieve the most pertinent information from the knowledge base, a task complicated by the inherent semantic gap between user queries in natural language and the way information is structured and stored within a knowledge base \citep{ma2023query}. This ``retrieval challenge" is a bottleneck for building effective RAG systems.

In this work, we tackle this problem with a new method called Query-Based Retrieval Augmented Generation (QB-RAG). Instead of matching a user's query directly to documents, QB-RAG first transforms the knowledge base into a high-quality set of answerable questions. The user's query is then matched to the most relevant pre-generated question to find the best context. Our core contribution is an automated, LLM-powered filter that vets each generated question for clarity and answerability, ensuring a reliable, query-centric knowledge base. This offline preparation is shown to significantly improve both retrieval and answer quality, providing a practical blueprint for more dependable RAG systems in healthcare.

\subsection*{Generalizable Insights about Machine Learning in the Context of Healthcare}
Our work offers several generalizable insights for developing, evaluating, and deploying Large Language Model (LLM) applications in safety-critical domain of healthcare:
\begin{itemize}
 \item \textbf{LLMs may Enable Automated, Quality-Controlled Curation of Knowledge Base for RAG:} LLMs can be employed to automatically generate diverse questions from trusted healthcare content and may subsequently act as an 'answerability filter' to ensure generated questions are relevant and verifiable against the source. This automated offline pipeline can create a high-quality, query-centric representation of knowledge base, enhancing grounding and reducing manual curation efforts often needed for reliable healthcare applications.
 \item \textbf{Comprehensive Evaluation Framework for Healthcare QA:} We introduce a comprehensive evaluation framework specifically designed for healthcare question-answering. This framework goes beyond standard retrieval metrics by incorporating measures of answer quality, including faithfulness, relevancy, and adherence to healthcare guidelines. This allows for a more nuanced assessment of how retrieval improvements translate into tangible benefits for downstream question-answering in a healthcare context.
 \item \textbf{RAG Performance is Sensitive to Offline Question Set Quality:}  Our findings indicate that the effectiveness of query-based RAG approach, particularly its ability to generalize to the variability of real-world user health queries, is correlated with the coverage and diversity of the pre-generated question set. Practitioners can optimize offline question generation to capture a wide range of answerable intents from a knowledge base for effective real-world performance based on their unique content properties.
 \item \textbf{Offline Query-Space Alignment Combines Low Inference Latency with Enhanced Retrieval:} Common RAG enhancement techniques involving multiple online LLM calls can introduce inference latency. Our work demonstrates that offline query-space alignment offers a practical solution. Pre-computing a query-centric knowledge representation can significantly reduce response times, making performant and reliable Healthcare-RAG systems more viable.
\end{itemize}
\section{Related Work}

\label{related_work}
To address this ``retrieval challenge", one set of approaches leverage the LLMs to create queries more semantically aligned with the knowledge base content. Query2Doc \citep{wang2023query2doc} and HyDE \citep{gao2022precise}, for example, generate a hypothetical document that would ideally answer the user's query. This synthetic document is then used as the retrieval key, improving the chances of finding relevant information. Similarly, QA-RAG \citep{kim2024rag} utilizes a fine-tuned LLM to generate a candidate answer, which is then used in conjunction with the original query to enhance retrieval. Another set of approaches diversifies search space by modifying user queries, such as duplicating or splitting queries and then using these queries to retrieve wider range of relevant documents \citep{ma2023query}. Methods like Dense Passage Retrieval \citep{karpukhin2020dense} focus on embedding fine-tuning, aiming to create semantically aligned representations of queries and documents, hoping to enhance retrieval precision.  Other work has focused on structuring the knowledge base itself, for instance by using knowledge graphs. Xu et al. \citep{xu2024} construct a graph from documents and retrieve relevant subgraphs to augment the generation process in customer service contexts.

Offline synthesis of pseudo-queries has also been explored for improving information retrieval. In doc2query and its iteration docTTTTTquery, \citep{nogueira2019documentexpansionqueryprediction, nogueira2019doc2query} showed that generating multiple queries from a document and concatenating them alongside the original document improves retrieval performance by enriching the document representation. More recently, \citep{raina2024questionbasedretrievalusingatomic} proposed a ``Question-Based Retrieval using Atomic Units" method for enterprise RAG applications. They decompose knowledge into question-answer pairs based on atomic units of information, advocating for a question-centric retrieval approach. Concurrent to our work, QuIM-RAG \citep{saha2024} also generates a representative question for each document, performing retrieval by finding the closest match in an "inverted question" index. Although these approaches share a similar high-level goal, they do not present scalable and robust query base filtering mechanisms, or they lack in comprehensive metrics at all levels of knowledge base curation, retrieval and answer generation, which are crucial for performance and error tracing in high stakes domains like healthcare.  
Our proposed method query-based RAG (QB-RAG) shares the spirit of these offline query generation approaches, but differentiates itself by focusing on generating answerable questions tailored for healthcare question-answering. This emphasis on answerability enforced by our LLM-based filtering mechanism is a crucial distinction, particularly in healthcare, where verifiable answers are paramount.

\section{Health Application and Dataset}
\label{sec:dataset}
Digital health programs represent a paradigm shift in managing chronic conditions such as Type 2 diabetes (T2D) and Hypertension (HTN). They can improve health markers associated with chronic conditions by delivering personalized care and support directly to patients \cite{majithia2020glycemic}. An integral component of these digital health programs is the messaging platform. However, the increasing demand for immediate healthcare information often strains the capacity of human healthcare providers to deliver timely support through messaging.

\begin{figure}[t]
\caption{Illustration of Content Cards in our mobile application, these cover broad range of topics, from operational health management to nutrition and physical advice.}
\centering
\includegraphics[scale=.4]{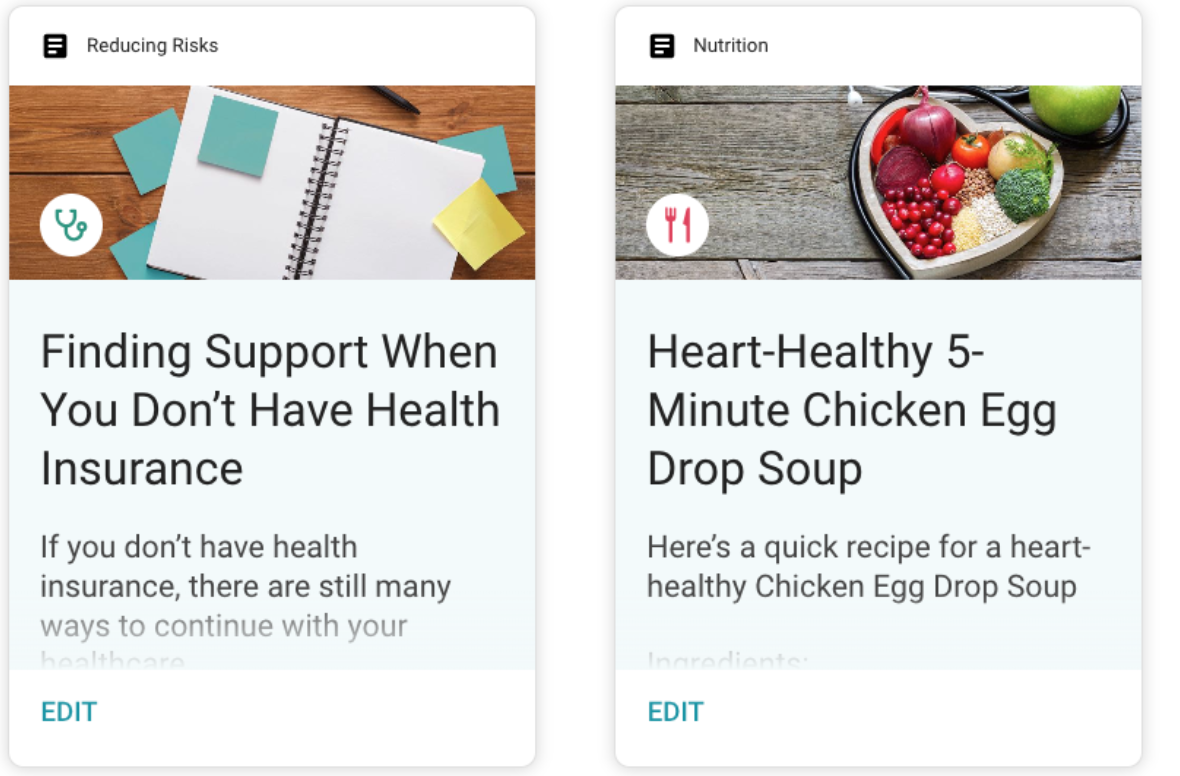}
\label{fig:onduo}
\end{figure}

To overcome this scalability challenge, digital health platforms are increasingly turning to meticulously curated content repositories designed to provide readily available answers to common patient questions. This study utilizes such a content repository, referred to as ``Content Cards,'' developed to address frequently asked questions (FAQs) from patients with T2D and HTN participating in our proprietary digital health programs. This dataset, encompassing 630 English-language content cards, covers a comprehensive range of topics pertinent to managing T2D and HTN. This includes general health information, detailed guidance on using platform-specific features and connected devices (e.g., blood glucose monitors), and personalized dietary recommendations. Figure \ref{fig:onduo} illustrates how these Content Cards are presented within our mobile application, while the excerpt in Appendix \ref{appendix:cc_example} provides a representative sample of the content.

\section{Mathematical Formulation}
\label{theory}
This section presents a mathematical framework for analyzing retrieval within RAG systems, highlighting the limitations of conventional approaches and motivating our proposed QB-RAG method.

\subsection{Notation}
Consider the set of $M$ content documents  $\mathcal{C} = \{c_1, \cdots, c_M\}$ and the set of $N$ questions $\mathcal{Q} = \{q_1, \cdots, q_N\}$ generated from $\mathcal{C}$ as specified in \ref{offline_q}. Unless otherwise specified, $c \in \mathcal{C}$ and $q \in \mathcal{Q}$ will refer to their respective embedding representations. We will assume that all embeddings are normalized and the distance function is the cosine distance denoted as $d(x,y)$.

The question generation process in section \ref{offline_q} is a one to many process, that is from one content we generate multiple questions. To encode this relationship, we denote the matrix $A \in \{0,1\}^{M \times N}$ where  $A_{ij} = \mathds{1} [c_i \text{ generated } q_j]$, where $\mathds{1}$ denotes the indicator function. Per our generation process and the relevance evaluation, it is understood that $A_{ij}=1 \Rightarrow q_j \text{ can be answered using } c_i$. However, the converse is not necessarily true as potentially different contents may be able to answer the same question. Thus, we define $A^* \in \{0,1\}^{M \times N}$, the ground-truth answerability matrix, which is the dense unobserved matrix such that $A^*_{ij} = \mathds{1} [q_j \text{ can be answered using } c_i]$. This implies that $A$ is a partial observation of $A^*$, which we call the oracle matrix. Specifically, $A_{ij} = 1$ if $A^*_{ij} = 1$ and $q_j$ was generated from $c_i$ during the offline question generation process; otherwise, $A_{ij} = 0$.
We point out that following our question generation process, we have $\forall c_j \in \mathcal{C}: \: \exists q_i \in \mathcal{Q} \text{ s.t. } A_{ij} = 1$.

For a new user question $q_0$, traditional RAG systems retrieve content that maximize some measure of similarity $\arg \max_{c \in \mathcal{C}} {c^t q_0}$ (or equivalently $\arg \min_{c \in \mathcal{C}} d(c, q_0)$).

For simplicity in our mathematical formulation, we will assume retrieval to mean a single piece of content in this section, where RAG retrieves the most similar content to the user query. 
We use Google's \texttt{textembedding-gecko} in the current work for generating embeddings. We denote the embedding matrices for content base and questions, respectively, with capital letters $C \in \mathbb{R}^{d \times M}$ and $Q \in \mathbb{R}^{d \times N}$ where $d$=768 is the dimension of our chosen embedder.

As a general note, we usually denote sets as calligraphic, vectors as lower case, matrices as upper case, and numbers/indices lower or upper case.

\subsection{Ideal Retrieval Objective}
Ideally during the retrieval phase, for a new user query $q_0$, one would evaluate all documents $c \in \mathcal{C}$ by asking the retrieval question {``Can the query $q_0$ be answered using the content $c$?''}. We call this the \textit{retrieval task} and define the \textit{retrieval function} $f^*(c, q_0) \in \{0,1\}$. Note this is the exact formulation of the matrix $A^*$ for the generated questions, i.e. $A^* = (f^*(c_i, q_j))_{ij}$.

Rather than pursuing this exact operation for every entry, many RAG systems typically approximate the retrieval task with some proxy as briefly described in \ref{related_work}. We briefly summarize them here in terms of our mathematical formulation.
\begin{itemize}
    \item \textbf{Naive RAG:} Approximates $f^*(c,q_0) \approx c^t q_0$.  Relies on cosine similarity between query and content embeddings, which can suffer from semantic misalignment. See a failing example in Appendix \ref{sec:failing-naive-rage}.

    \item \textbf{LLM-Rewriting Methods (HyDE, Query2Doc, QA-RAG):}  Approximate \\ $f^*(c, q_0) \approx c^t p_{LLM}(q_0)$, using LLMs to rewrite queries to better align with content space. Here, $p_{LLM}(q_0)$ refers to the LLM’s output conditioned on the original query aligned with the content space. The effectiveness of this approach is highly dependent on the quality of LLM's generation and alignment with the content base embeddings.
    \item \textbf{Adapter/Fine-tuning Methods:}  Approximate $f^*(c,q_0)$ by learning transformations or specialized embeddings to improve alignment.  These often require expensive training data and may not generalize well to different datasets.
\end{itemize}

\subsection{QB-RAG}
\label{theory-qbrag}
QB-RAG leverages a pre-computed set of questions, $\mathcal{Q}$, derived from the content base $\mathcal{C}$. These questions are generated offline as described in Section \ref{offline_q}, mitigating concerns about online computational overhead. In contrast to methods relying on online LLM calls for query rewriting, QB-RAG shifts this computational burden offline. This distinction is significant, as online rewriting necessitates serial LLM invocation, potentially introducing latency detrimental to user experience. Furthermore, QB-RAG's direct alignment within the query space offers a more transparent and interpretable retrieval process compared to the implicit alignment strategies of LLM-based rewriting techniques.

\subsubsection{Vanilla QB-RAG}
Our vanilla approach of QB-RAG first generates an extensive set of questions that are known to be answered by the content by initializing $\mathcal{Q}$ and $A$. This operation can happen offline and upon uploading new documents to our content base.
Second, for an online query, QB-RAG searches for similar question (resp. questions) within $\mathcal{Q}$ by finding $\arg \min_\mathcal{Q} d(q_0, q)$ (resp. $\arg \min_{\mathcal{S} \subset \mathcal{Q}} \sum_{q \in \mathcal{S}}{d(q, q_0)}$).
Given we are now comparing questions to questions, we expect the distance measure to be calibrated given the comparison is also aligned.
Third upon retrieving similar questions, the associated contents are fetched, and fed to the generative LLM like other RAG systems (after dropping duplicate contents if necessary). We provide the full details in Algorithm \ref{alg:vanill-qb-rag}.

\begin{algorithm}
\caption{Vanilla QB-RAG}\label{alg:vanill-qb-rag}
\begin{algorithmic}
    \REQUIRE $\mathcal{C}, \mathcal{Q}, A$ (or $A^*,\hat A$), new query $q_0$, target number of retrievals $k \leq M$
    \STATE Similarities $z \gets Q^t q_0 \in \mathbb{R}^N$
    \STATE Sort $z$ by descending values and sort $\mathcal{Q} \& A$'s columns accordingly.
    \STATE $\mathcal{S} \gets \{\}$
    \FOR{$j \in [1..N]$}
        \STATE $i \gets i$ s.t. $ A_{ij} = 1$ identify the content associated associated to the $j$-th question.\\
        \textit{\scriptsize If $A$ is non-sparse, we break ties by favoring new content, by number of associated questions, then at random.}
        \STATE $\mathcal{S} \gets \mathcal{S} \cup \{ c_i \} $ \textbf{if} $c_i \notin \mathcal{S}$
        \STATE \textbf{If} {$\lvert \mathcal{S} \rvert = k$} \textbf{then break}
    \ENDFOR
    \STATE \textbf{Return} $\mathcal{S}$
\end{algorithmic}
\end{algorithm}

\subsubsection{Oracle and Approximate QB-RAG}
In this section, we present oracle and approximate variants as conceptual extensions. We leave experimental evaluation of these variants as future work.

We turn our attention to the best possible retrieval system using a query based alignment and retriever. We previously noted that the matrix $A$ indicating which content generated which question, is actually a \textbf{sparse partial observation} of $A^*$ indicating which content is relevant to a question. For this section we will assume that we have access to the oracle matrix $A^*$ or some approximation of it denoted $\hat{A}$. For simplicity, we describe the algorithms with ${A}^*$, but is also applicable to $\hat{A}$.

Before getting into the algorithmic details, let us first motivate the use of this matrix $A^*$. We note that while computing the matrix completely would involve $N \times M$ LLM calls, these can all happen offline. Further depending on the dataset at hand, this may not even prohibitive in cost. E.g. in our dataset, this would lead to more than 5000 offline calls. Let us point out though that exactly computing $A^*$ is not necessary, and some approximation of it could also lead to improved retrieval. In fact, we propose two ways of computing the estimate $\hat{A}$:
\begin{enumerate}
    \item Only compute matrix entries $A^*_{ij}$ such that $c_i^t q_j \geq \lambda$ where lambda is some threshold, which can be chosen in order to compute only some percentile of entries. This typically filters out combinations of contents and queries that are likely not relevant to each other.
    \item After computing a set of entries (whether chosen at random or according to the above rule), we can rely on matrix completion techniques to infer and approximate the remainder of the matrix. Matrix completion is highly efficient on moderate to large matrix sizes.
\end{enumerate}

Given the oracle $A^*$ or some estimate $\hat{A}$ thereof, we can adapt Algorithm \ref{alg:vanill-qb-rag} to incorporate the non-sparse nature of $A^*$ or $\hat{A}$. Specifically, since multiple content pieces may be associated with a single question, we introduce tie-breaking rules within the algorithm that first prioritize newer content (especially relevant to healthcare care). We then prioritize the content capable of answering a diverse range of questions within $\mathcal{Q}$. This reflects our intuition that such content is likely to be more broadly relevant.


\section{Methods}
\subsection{Offline Question Generation}\label{offline_q}
The core premise of our work is to improve content retrieval in RAG systems by directly aligning incoming user queries with a pre-computed set of questions derived from our content base. This requires generating a comprehensive set of questions that are answerable by our content, which we achieve through a two-step process. All LLM generations utilize \texttt{Gemini\--Pro} and text embeddings are generated using \texttt{textembedding\--gecko} – both of which are commercially available from Google.

\subsubsection{Base prompt}
To generate an initial set of questions, we design a prompt (Appendix \ref{appendix:prompts}.\ref{lst:base-prompt}) with instructions and few-shot examples. This prompt incorporates a \texttt{num\_questions} parameter to specify the target number of questions per content card. While we set this parameter to 20, the actual number of questions generated by the LLM may vary. Applying this prompt to our content base of 630 cards resulted in over 8,000 potential questions.

\subsubsection{Answerability model}
\label{sec:answerability}
To mitigate issue of irrelevancy of the generated questions to the content, we developed an LLM-based answerability model. For each content card and generated question pair, we prompt the LLM to assess the content's ability to answer the question. We structure the prompt to elicit a step-by-step explanation of the answerability judgment (Appendix \ref{prompt:answerability}). 

This filtering process resulted in a refined set of approximately 4,800 answerable questions derived from our 630 content cards. Each question in this curated set is directly mapped back to its source content card, which serves as the ``golden'' source for a correct answer. 

To validate the effectiveness and reliability of our answerability model, we randomly selected 100 question and content card pairs and evaluated them using both our automated model and three clinical experts. The clinical experts rated the answerability of each pair on a three-point scale: ``Content answers completely,'' ``Content answers partially,'' or ``Content does not answer.'' The results demonstrated strong agreement between our model's assessments and the clinical expert's judgments. 90\% of the pairs received the same answerability rating, with an additional 9\% categorized as ``partially answerable,'' indicating some degree of inherent subjectivity in the evaluation task. To further validate the answerability model, we additionally confirmed that the semantic similarity among the generated questions above are as expected (Appendix \ref{gen_q_similarity}). This comprehensive analysis, coupled with the increasing recognition of LLMs as auto-evaluation tools \cite{lee2023rlaif}, supports the validity of our answerability model for filtering irrelevant questions. 

\subsection{Benchmarks}

This section outlines the retrieval methods evaluated in our benchmark. To ensure a fair comparison, each method retrieves the same number of documents, denoted by $k$, for a given query (the value of $k$ is varied across experiments). We evaluate three methods - Naive RAG \citep{lewis2020retrieval}, HyDE\citep{gao2022precise}, QA-RAG\citep{kim2024rag} - against our Vanilla QB-RAG approach. The details of the three benchmark methods are described in Appendix \ref{benchmark-methods}. We run the simplest version of our method (see Algorithm \ref{alg:vanill-qb-rag}).
Given this method is highly dependent on the questions generated offline, we parameterize {QB-RAG}-${\bar m}$ where $\bar m$ measures the average number of questions generated per content. $\bar m$ reflects the ``coverage'' extracted from each content, implying that higher coverage should lead to improved retrieval quality, and consequently, better-generated answers. We decrease $\bar m$ by down-sampling the set of questions in our experiments, to effectively reduce the coverage of our question base. The maximum value from our generation is $\bar m = 8$ which includes our entire question set.

We note that that the methods presented above are not mutually exclusive and can be combined to potentially achieve improved performance, as illustrated by {QA-RAG} \citep{kim2024rag}. Our experiments focus on evaluating the efficacy of each method in isolation.


\subsection{Answer Generation}
After retrieving relevant content using the methods described above, we employ an LLM (\texttt{Gemini-Pro}) for answer generation. For each question, the retrieved content is provided as context to the LLM along with the prompt in Appendix \ref{prompt:answer} to generate the answer.

\subsection{Test Data Generation}
We construct two distinct test sets designed to assess the performance of the various retrieval methods under different conditions. Each test set consists of questions answerable by our content base, ensuring relevance to the task. Crucially, no test question appears verbatim within the pre-generated question set used for retrieval. All test questions are generated using \texttt{Gemini-Pro}.
\begin{enumerate}
    \item \texttt{Rephrase}: To generate this test set, we prompted an LLM to rephrase each of the 4.8k questions in the knowledge base. Then, 500 of those were randomly sampled to yield the first test set. This approach was done to simulate scenarios where the knowledge base contains a comprehensive set of questions. In such scenarios, the intents of new incoming questions are more likely to be represented in the knowledge base, as might occur for mature systems.
    
    \item \texttt{Out-of-Distribution}: To generate the second test set, we prompted an LLM to generate a new question for each of the 630 contents. The LLM was instructed to not generate a question that already exists in the question knowledge base for that content. Then, the 630 newly generated questions were filtered via the answerability model to ensure the new questions were indeed answerable by the corresponding content. After filtering, 305 newly generated questions made up the second test set. While questions in this Out-of-Distribution test set do not inquire the same exact knowledge as the existing questions in the retrieval pool, they are related and can be connected to an existing question in the embedding space. This test set reflects the adversarial, low-coverage scenarios (e.g.  systems with a relatively cold start). 
\end{enumerate}


\section{Metrics}
\label{sec:metrics}
To rigorously assess the performance of our proposed QB-RAG method for healthcare applications, we employ two distinct sets of metrics: those evaluating the quality of content retrieval and those assessing the quality of the generated answers. While QB-RAG's primary objective is to enhance retrieval, we hypothesize that this improvement will translate to better-generated answers.

\subsection{Retrieval Evaluation}
\label{retrieval_eval}
\label{sec:metrics-retrieval}
\begin{itemize}
\item \textbf{Exact Recovery Rate}: This metric measures the percentage of test questions for which the retrieved set of $k$ documents includes the exact content piece used to generate the original question. We acknowledge that while exact recovery represents retrieval of ``golden content'', multiple contents might offer relevant information for a single question. Therefore, we incorporate additional relevance measures described below.

\item \textbf{Auto-evaluator Relevancy Rate}: This metric addresses the limitations of relying solely on exact matches by leveraging the answerability model to gauge content relevance. It calculates the percentage of test questions for which at least one retrieved document is deemed relevant by the answerability model. This automated assessment has demonstrated strong correlation with human judgments as as described previously.
\end{itemize}

\subsection{Answer Evaluation}\label{answer_eval}

\begin{itemize}
\item \textbf{Answer Guideline Adherence Rate:} The assessment involves a three-step process. Initially, an LLM generates a ``golden answer'' for each test question using the associated content. Subsequently, another LLM analyzes this golden answer to create a guideline outlining the key elements an accurate response should contain. Finally, a third LLM evaluates the candidate answer against this extracted guideline, assigning a score from 0 to 1 based on the extent to which the answer covers the guideline's key points. While this guideline-based approach aims to capture the nuances of answer quality, it relies on the accuracy of both the golden answer and the extracted rubric. To provide a more robust and multifaceted evaluation, we introduce additional metrics that directly assess distinct aspects of answer quality below.

\item \textbf{Answer Relevancy Rate}: This metric evaluates whether a generated answer directly address the user's question and provides a self-contained response. An LLM classifies each answer as either relevant (YES) or not relevant (NO) to the corresponding test question, focusing solely on the relevance, not the accuracy or factual correctness of the answer.

\item \textbf{Answer Faithfulness Rate}: This evaluates whether the content supplied to the LLM during generation supports the generated answer. An answer is deemed faithful if \textit{any} portion of the provided content supports it, regardless of the presence of irrelevant content. This component is crucial to ensure that the answer generated is grounded on our content, a crucial feature for many healthcare applications. We penalize any unwarranted extrapolation or information not grounded in the provided content. Note that declined answers are considered not faithful in our definition.

\item \textbf{Answer Declined Rate}: This metric assesses whether the LLM declined to answer the question. This may occur because our answer generation couldn't find relevant information in the content (we specifically prompt the LLM to only answer questions when there is a retrieved content.)
\end{itemize}

\subsection{Statistical Analysis}
To validate the performance differences between models, we conducted \textbf{statistical significance testing}, comparing each model against the NaiveRAG baseline. For binary evaluation metrics, we used a one-sided exact binomial test on the discordant pairs. For the continuous guideline adherence rate, we employed the one-sided Wilcoxon signed-rank test. For all analyses, a p-value of p\textless0.05 was considered statistically significant.

\subsection{Note on Auto-Evaluator Use}
For all metrics that require LLMs for annotation (all but exact recovery), we conducted manual annotation to validate the use of auto-evaluators. In section \ref{sec:answerability}, we separately validated the auto-evaluator relevancy for retrieval on 100 annotations. For answer related metrics, we conducted 2 manual annotations on 50 questions each, ensuring diverse sets, with full results in Appendix \ref{sec:answer-qual-annot}.
Altogether the annotation showed 90+\% agreement between auto-evaluators and human annotation for binary tasks, and high correlation (.84 Spearman rank correlation) on scoring tasks. Inter-rater agreement is comparable, validating the use of auto-evaluators throughout the experiments.


\section{Results}
\label{sec:exp}
In this section we discuss the performance of our methods, QB-RAG-8 (resp. QB-RAG-2) where the knowledge base has an average of 8 (resp. 2) queries per content, which are compared against the incoming query for retrieval. We assess the impact of QB-RAG on both retrieval efficacy and the quality of generated answers.

\subsection{Effect of QB-RAG on Retrieval Efficacy}
On the \texttt{Rephrase} test set, where the knowledge base is expected to contain questions semantically similar to the test questions, QB-RAG-8 consistently outperforms all benchmark methods.  As shown in table \ref{tab:exp_rephrase}, QB-RAG-8 nearly doubles the exact recovery rate compared to other methods (e.g., from 45\% to 89\% when retrieving a single document). This suggests that a comprehensive, query-aligned knowledge base, as constructed by QB-RAG-8, substantially improves the retrieval of the exact source document. These impressive gains highlight the scenario where our knowledge base is comprehensive and covers quite broadly the extent of questions our documents can answer. 

The exact recovery rate does not present the full picture. When analyzing the Auto-evaluator Relevancy (refer section \ref{retrieval_eval}), our method still outperforms the baselines when retrieving a single content. As measured by the LLM, the content we retrieve are relevant 68\% of the time, whereas traditional methods retrieve relevant content around 40\% of the time.

Increasing the number of retrieved documents to 3 further illustrates the efficacy of QB-RAG-8. The exact recovery rate reaches a near-optimal 97\%, substantially higher than the baseline approaches. QB-RAG-8 also achieves almost 20\% higher relevancy rate compared to the baselines. These findings underscore the robustness of QB-RAG-8 in retrieving both the target document and a set of relevant documents.

\begin{table}[t]
    
\small
\centering
\setlength{\tabcolsep}{4pt}  
\scriptsize
\begin{tabular}{lcccccc}
\toprule
\makecell{\textbf{Retrieval assessment} \\ \texttt{Rephrase}} & \multicolumn{2}{c}{\makecell{Exact Recovery \\ Rate}} &
\multicolumn{2}{c}{\makecell{Auto-evaluator \\ Relevancy Rate}} \\
\cmidrule(r){1-1}\cmidrule(r){2-3} \cmidrule(r){4-5} \cmidrule(r){6-7}
Number of retrieved docs & 1 & 3 & 1 & 3  \\
\midrule
QB-RAG-8 & \textbf{0.89}* & \textbf{0.97}* & \textbf{0.68}* & \textbf{0.76}* \\
QB-RAG-2 & 0.59* & 0.75* & 0.53* & 0.66*  \\
Naive RAG & 0.44 & 0.61 & 0.39 & 0.56  \\
QA-RAG & 0.45 & 0.60 & 0.41 & 0.56  \\
HyDE & 0.47 & 0.63 & 0.41 & 0.56  \\
\bottomrule
\end{tabular}
\caption{Retrieval performance (higher is better) of methods on \texttt{Rephrase}, where documents were retrieved given rephrased questions from the content base.\textit{* denotes statistical sign.}}

    \label{tab:exp_rephrase}
\end{table}

On the more challenging \texttt{Out-of-Distribution} test set, QB-RAG-8 maintains its advantage, albeit with smaller gains. Table \ref{tab:exp_ood} shows that QB-RAG-8 improves the exact recovery rate by 1.3\% to 6.2\% compared to benchmark methods. Despite the adversarial nature of this test set, where incoming questions are intentionally dissimilar to the training set, QB-RAG-8 consistently retrieves the correct source document more often. 

Furthermore, QB-RAG-8 demonstrates a significant improvement in relevancy. The auto-evaluator relevancy rate shows gains of 8\% to 15\% over baseline methods, indicating that QB-RAG-8 effectively identifies relevant content even when the query distribution shifts. These improvements in retrieval quality directly benefit downstream answer generation, as more relevant content is likely to lead to more accurate and informative answers. 
\begin{table}[t]
        \small
    \centering
    \setlength{\tabcolsep}{4pt}  
    \scriptsize
    \begin{tabular}{lcccccc}
    \toprule
    \makecell{\textbf{Retrieval assessment} \\ \texttt{Out-of-Dist}} & \multicolumn{2}{c}{\makecell{Exact Recovery \\ Rate}} & \multicolumn{2}{c}{\makecell{Auto-evaluator \\ Relevancy Rate}} \\
    \cmidrule(r){1-1}\cmidrule(r){2-3} \cmidrule(r){4-5} \cmidrule(r){6-7}
    Number of retrieved docs & 1 & 3 & 1 & 3  \\
    \midrule
    QB-RAG-8 & \textbf{0.53} & \textbf{0.72} & \textbf{0.58}* & \textbf{0.75}* \\
    QB-RAG-2 & 0.42 & 0.57 & 0.50 & 0.64 \\
    Naive RAG & 0.50 & 0.68 & 0.50 & 0.64 \\
    QA-RAG & 0.51 & 0.66 & 0.47 & 0.60 \\
    HyDE & 0.52 & 0.70 & 0.48 & 0.65 \\
    \bottomrule
    \end{tabular}
    \caption{Retrieval performance (higher is better) of methods on \texttt{Out-of-Distribution}, where documents were retrieved given newly generated questions.\textit{* denotes statistical sign}.}

    \label{tab:exp_ood}
\end{table}

\subsection{Effect of QB-RAG on Generated Answer Quality}
We now examine how the improved retrieval accuracy of QB-RAG translates to the quality of generated answers. Recall that the \texttt{Rephrase} test set favors our query-based approach as test questions are semantically similar to those in our generated knowledge base.

As shown in Table \ref{tab:exp_reph_ans}, both QB-RAG-8 and QB-RAG-2 consistently outperform the benchmark methods on all answer quality metrics for the \texttt{Rephrase} test set. Notably, QB-RAG-8 achieves an 84\% answer faithfulness rate, significantly surpassing the 62\%-68\% rates of the baseline methods. This suggests that by accurately retrieving the most relevant content, QB-RAG enables the LLM to generate answers that are well-grounded in the provided information. Additionally, QB-RAG achieves the highest answer guideline adherence rate, indicating its answers effectively address the key elements outlined in the pre-defined guidelines (refer to Section \ref{answer_eval}).

\begin{table}[t]

    \small
    \centering
    \setlength{\tabcolsep}{4pt}  
    \scriptsize
    \begin{tabular}{lcccc}
    \toprule
    \makecell{\textbf{Answer assessment} \\ \texttt{Rephrase}} & \makecell{Declined \\ Rate} & \makecell{Faithfulness \\ Rate} & \makecell{Answer \\ Relevancy Rate} & \makecell{Guideline \\ Adherence Rate} \\
    \midrule
    QB-RAG-8 & \textbf{0.12}* & \textbf{0.84}* & \textbf{0.83}* & \textbf{0.79}* \\
    QB-RAG-2 & 0.21* & 0.74* & 0.73* & 0.73* \\
    Naive RAG & 0.30 & 0.67 & 0.66 & 0.68 \\
    QA-RAG & 0.33 & 0.62 & 0.62 & 0.62 \\
    HyDE & 0.30 & 0.68 & 0.66 & 0.67 \\
    \bottomrule
    \end{tabular}
    \caption{Answer quality of methods on \texttt{Rephrase}, where documents were retrieved given rephrased questions from the content base. 3 documents retrieved.\textit{* denotes statistical sign.}}

    \label{tab:exp_reph_ans}
\end{table}

These patterns hold on the more challenging \texttt{Out-of-Distri\-bution} test set, as seen in table \ref{tab:exp_ood_ans}.
There QB-RAG-8 achieves 78\% faithfulness, fairing higher than the benchmarks 68\%-74\%. However QB-RAG-2 under-performs, highlighting its inability to generalize to new questions.

\begin{table}[t]
     
    \small
    \centering
    \setlength{\tabcolsep}{4pt}  
    \scriptsize
    \begin{tabular}{lcccc}
    \toprule
    \makecell{\textbf{Answer assessment} \\ \texttt{Out-of-Dist}} & \makecell{Declined \\ Rate} & \makecell{Faithfulness \\ Rate} & \makecell{Answer \\ Relevancy Rate} & \makecell{Guideline Adherence \\ Rate} \\
    \midrule
    QB-RAG-8 & \textbf{0.17}* & \textbf{0.78}* & \textbf{0.77}* & \textbf{0.69} \\
    QB-RAG-2 & 0.29 & 0.66 & 0.65 & 0.63 \\
    Naive RAG & 0.27 & 0.72 & 0.70 & 0.64 \\
    QA-RAG & 0.29 & 0.68 & 0.67 & 0.63 \\
    HyDE & 0.24* & 0.74* & 0.72* & 0.68 \\
    \bottomrule
    \end{tabular}
    \caption{Answer quality of methods on \texttt{Out-of-Distribution}, where documents were retrieved given newly generated questions. 3 documents retrieved.\textit{* denotes statistical sign.}}

    \label{tab:exp_ood_ans}
\end{table}
An answer can be deemed unfaithful either because the answer is not grounded, or because the LLM declined to answer. We note that the higher faithfulness rate of QB-RAG-8 is largely due to improved retrieval, resulting in a decline rate of only 12\% on the \texttt{Rephrase} test set and 17\% on the \texttt{Out-of-Distribution} test set, compared to significantly higher rates for the baselines. We observe a slight increase in answers that are not grounded, from $\approx2\%-3\%$ on the \textit{Rephrase} test set to $\approx3\%-5\%$ on the \texttt{Out-of-Distribution} test set for QB-RAG. Importantly, we achieve the lowest unfaithfulness rate (3\%) on the \texttt{Rephrase} test set with QB-RAG-8, further underscoring the value of generating a comprehensive question set for optimal coverage.

While QB-RAG significantly improves answer faithfulness through better retrieval, it's worth noting that the groundedness aspect of faithfulness rate is ultimately determined by the capabilities of the answer generation module itself. Further enhancements to answer faithfulness could involve techniques like preference modeling and RLHF where the LLM is specifically for this objective  \citep{bai2022training}.

\subsection{Effect of Coverage of Generated Questions}

To examine the relationship between the breadth of the generated question set and QB-RAG's effectiveness, we compare the performance of QB-RAG-2 and QB-RAG-8. This analysis reveals a strong dependence on question coverage.

On the \texttt{Rephrase} test set, QB-RAG-2, despite its reduced question set, still surpasses other retrieval methods (table \ref{tab:exp_rephrase}). However, the magnitude of improvement is noticeably smaller compared to QB-RAG-8. For instance, QB-RAG-2 shows a 10-15\% improvement in exact recovery rate over baselines, whereas QB-RAG-8 nearly doubles the exact recovery rate. This pattern also holds for relevancy rate, indicating that a larger, more comprehensive question set translates to more effective retrieval.

The \texttt{Out-of-Distribution} test set (table \ref{tab:exp_ood}) reveals a more pronounced impact of question base coverage. Here, QB-RAG-2's performance drops below that of some benchmark methods, with the exact recovery rate decreasing by 7-12\%. Interestingly, QB-RAG-2 still achieves comparable performance on relevancy-based metrics, suggesting that even a limited question base can partially capture relevant content, but may not pinpoint the exact source document as effectively.

This sensitivity analysis clearly shows that the performance of QB-RAG methods are very much tied to our ability to generate a comprehensive set of questions from RAG knowledge base. As we increase the number of generated questions in our database, we expect to see better retrieval, higher exact matches and higher relevance. Beyond simply increasing the number of questions, improving their diversity may be another way for further improving QB-RAG's performance. By generating a more diverse set of questions for each content piece, we can capture a broader range of semantic nuances and user intents. This is particularly relevant given that modern retrieval algorithms efficiently handle large knowledge bases, making quantity less of a limiting factor in doing efficient retrieval.

This sensitivity to question coverage is further evident in the quality of the generated answers (Tables \ref{tab:exp_reph_ans} and \ref{tab:exp_ood_ans}). QB-RAG-8 consistently leads to more faithful, relevant, and accurate answers compared to QB-RAG-2, directly reflecting the differences observed in their retrieval performance. These findings highlight that QB-RAG's success in downstream tasks is fundamentally linked to its ability to construct and leverage a comprehensive and diverse question set that effectively captures the content and semantic nuances of incoming queries within the RAG knowledge base.

\subsection{MedQUAD Experiment}
We additionally compared the methods with the MedQUAD dataset,
which includes pairs of human-authored questions with supporting sources, across diverse healthcare topics \citep{BenAbacha-BMC-2019}. We provide key results in Appendix \ref{medquad-results}, showing up to 20\% gain in retrieval accuracy and 10\% for final answer quality for select metrics.


\section{Discussion} 

This paper introduces QB-RAG, a framework for enhancing the retrieval phase of RAG systems, with a particular focus on healthcare question-answering. QB-RAG tackles the challenge of semantic misalignment between user queries and knowledge bases by pre-generating a comprehensive set of answerable questions directly from the healthcare content. While the QB-RAG approach shares similarities with prior work on offline query synthesis for information retrieval, its emphasis on generating answerable questions tailored for healthcare, combined with the LLM-based answerability filtering, distinguishes it within the broader landscape of RAG system enhancement techniques. Unlike conventional methods that rely on online LLM calls for query rewriting or enhancement, QB-RAG shifts the computational burden of question generation upfront and offline, thereby reducing latency at inference time. Furthermore, QB-RAG benefits from the efficiency of highly optimized retrieval algorithms, even when handling large question sets, ensuring rapid and practically tractable content retrieval for incoming online queries.

Our experiments demonstrate that QB-RAG outperforms benchmark methods on two distinct test sets: the \texttt{Rephrase} set, which simulates a scenario with comprehensive question generation, and the more challenging \texttt{Out-of-Distribution} set, designed to assess robustness under varying conditions. The results highlight QB-RAG's ability to improve not only the exact recovery and relevance of retrieved documents but also the quality of downstream answer generation, as evidenced by higher scores on faithfulness, relevancy, and guideline adherence metrics. Crucially, answers generated using QB-RAG are more frequently grounded in the trusted source content, which is important for delivering reliable, up-to-date information to patient queries. Our sensitivity analysis further reveals that the effectiveness of QB-RAG is tied to the coverage and diversity of the generated question set. This suggests that future work could explore more sophisticated prompt engineering, supervised fine-tuning, or RLHF frameworks to enhance the diversity of generated questions. Metrics like BERTScore \citep{zhang2019bertscore} and Word Mover's Distance \citep{kusner2015word} could be used to evaluate this diversity, and an LLM could serve as an auto-evaluator in this regard.

QB-RAG's strengths extend beyond direct health question-answering, and can be valuable when integrated into EHR systems for telemedecine platforms with appropriate considerations of real-world implications. Several aspects, such as complying with relevant regulations, gauging the need for human supervision in for high-risk queries, and determining ethical deployment paths that track with the system’s readiness for different risk levels (as they relate to misinformation or hallucinations) will be necessary For efficient and accurate functionality of QB-RAG in various healthcare settings.

Despite its strengths and promising results, our work has several limitations that warrant further investigation. First, our evaluation focused on a curated healthcare content base consisting of well-crafted, concise pieces. Assessing QB-RAG's performance on larger, more diverse datasets, including public datasets or those from other domains and containing less structured or longer-form content, remains an important area for future research. This will establish the generalizability of our findings beyond the specific context of this study. Future research could also expand the system’s capabilities to handle increasingly wider scopes of queries, including unanticipated queries. This could include integrating hybrid approaches with benchmark methods (Appendix {\ref{benchmark-methods}}), dynamically updating content base (and therefore question base) from trusted sources, or adding incoming queries incrementally to the deployed system (after matching and validation) to continuously expand the retrieval pool. 

Second, the current study relies on LLM-based evaluation metrics. While LLMs have shown promise in assessing answer quality, the real-world deployment of QB-RAG in healthcare settings necessitates thorough expert review for each use case, especially for patient-facing applications. Direct patient interactions demand the highest level of scrutiny to ensure accuracy, safety, and reliability of generated information. Therefore, while automated evaluation serves as a valuable first step, human expert validation remains paramount for clinical deployment.

Third, a broader challenge inherent to all RAG systems, particularly in fast-changing domains like healthcare, is the dependence on the quality and recency of the underlying knowledge base. Medical consensus and best practices are constantly evolving. Thus, content needs to be regularly reviewed and updated. Developing robust mechanisms for ongoing knowledge base maintenance and ensuring alignment with current clinical guidelines is crucial, regardless of the specific retrieval methods employed.

Fourth, while we've positioned the offline question generation as an advantage for real-time applications, it's important to acknowledge that the initial generation and filtering process can be computationally expensive. Future work could explore methods for optimizing this process, such as more efficient filtering mechanisms, more targeted question generation strategies, parallelization \mbox{\citep{zhou2024surveyefficientinferencelarge}}, active learning, smaller model distillation, or, as indicated above, progressively expanding the retrieval pool leveraging actual queries.


\bibliography{sample}

\newpage
\appendix
\label{sec:appendix}

\section{MedQUAD Results}
\label{medquad-results}
We filtered MedQuAD\footnote{\href{https://huggingface.co/datasets/lavita/MedQuAD}{https://huggingface.co/datasets/lavita/MedQuAD}} down to 1000 questions of the most prevalent disease areas, then removed 2 duplicates. We used the resulting 998 human derived questions as test queries, and the 114 associated unique websites as our document base.
For QB-RAG we generated queries, resulting in 3068 total (recent model versions were more verbose enabling larger knowledge base). We report key results for QB-RAG-27 and QB-RAG-8 by down-sampling.

For fairness across benchmarked methods, we used Gemini-2.0-flash (previous models are unavailable) for all LLM tasks. See table for full results.
\begin{enumerate}
    \item For 1 document retrieved, QB-RAG-27 and QB-RAG-8 achieve 50-52\% exact recovery, while baselines are 31-37\%. Document relevancy rate soared from 83-87\% to 90-94\%!
    \item By all metrics, final answer quality improved: Answer relevance (resp. Faithfulness; Guideline adherence) improved from 90-94\% (resp 92-95\%; 53-56\%) to 94-97\% (resp 96-98\%; 59-63\%) using QB-RAG-27 and QB-RAG-8
    \item For 3 documents retrieved, scores are similar across all methods, indicating less obvious benefit from QB-RAG in this specific setting
\end{enumerate}

\begin{table}[ht]
    \small
\centering
\setlength{\tabcolsep}{4pt}
\scriptsize
\begin{tabular}{lcccccc}
\toprule
\makecell{\texttt{MedQUAD}\\Human\\Questions} & \makecell{Exact \\ Recovery \\Rate} & \makecell{Auto-evaluator\\Relevancy Rate} & \makecell{Answer\\Declined\\Rate} & \makecell{Faithfulness\\Rate} & \makecell{Answer \\ Relevancy \\ Rate} & \makecell{Guideline \\ Adherence \\Rate} \\
\midrule
QB-RAG-27 & 0.50* & \textbf{0.94} & \textbf{0.01}* & \textbf{0.98}* & \textbf{0.97}* & \textbf{0.63}* \\
QB-RAG-8  & \textbf{0.52}* & 0.90 & 0.03* & 0.96* & 0.94* & 0.59* \\
NaiveRAG  & 0.31 & 0.83 & 0.06 & 0.92 & 0.90 & 0.53 \\
QA-RAG    & 0.34* & 0.86 & 0.03* & 0.94* & 0.91* & 0.54 \\
HyDE      & 0.37* & 0.87 & 0.03* & 0.95* & 0.94* & 0.56 \\
\bottomrule
\end{tabular}
\caption{Evaluation metrics for different models across Retrieval and Answer quality dimensions on the MedQUAD dataset. Report for 1 document retrieved. \textit{* denotes statistical sign.}}
\label{tab:model-evaluation-medquad}

    \label{tab:model-evaluation-medquad}
\end{table}

Altogether, we believe that QB-RAG can improve over SOTA RAG methods in many settings. Yet as seen using 3 documents, this depends on the dataset and hyper-parameters.
Nevertheless, combining QB-RAG with existing RAG techniques is promising future work, that would enhance all retrieval by leveraging the best capabilities of each method.

\section{Additional Validation Results of Auto-Evaluator}
\label{sec:answer-qual-annot}
For this annotation task, we first point out that retrieval metrics (section \ref{sec:metrics-retrieval}) are either rule based or already aligned with human labels (see review in section \ref{sec:answerability}).

Regarding answer evaluation, we constructed 2 separate annotations task to ensure the auto-evaluation (AE) is aligned. We had 2 domain experts annotate the 50 questions for each task (100 total test questions were manually annotated), while blinded to the AE. responses. We separated the tasks to balance out the underlying AE labels given there is class imbalance for each.
\begin{itemize}
    \item \textbf{Task 1}: For faithfulness, relevancy and declined-to-answer, we sample a balanced set of 50 auto-evaluated questions (with associated answers and retrieved documents) across our method and different benchmarks (see 5.2).
    Human evaluators showed 100\% agreement with AE on declined to answer (inter-rater agreement 100\%), 97\% agreement on answer relevancy (inter-rater agreement 98\%), 90\% agreement on faithfulness (inter-rater agreement 90\%).
    \item \textbf{Task 2}: For guideline adherence, we sample a set of 50 questions (with associated answer and guideline) with distributed AE scores, across our method and different benchmarks. The 2 domain experts scored the guideline adherence on a 5 point likert scale.
    Humans evaluators showed .84 Spearman rank correlation with AE (inter-rater correlation .92), and associated .80 Pearson correlation with AE (inter-rater correlation .95).
\end{itemize}

There were no noticeable differences across methods. Taken together, the above results prove \textbf{AE is well calibrated} (i.e. no more disagreement than within experts) for binary tasks and \textbf{preserves ranking for quantitative tasks} with high correlation. This section boosts are confidence in using AE for assessing retrieval performance and final answer quality.

\section{Illustrating Semantic Misalignment on Naive RAG}
\label{sec:failing-naive-rage}
Here is a failing example for Naive RAG illustrating semantic misalignment between question and document space.
\begin{itemize}
        \item $q_1: $ What are benefits of soluble and insoluble fiber?
        \item $q_2: $ What are examples of soluble and insoluble fiber?
         \item $c_1: $ Soluble fiber helps lower cholesterol and keeps blood glucose from rising quickly. Insoluble fiber helps move food through your body, which can keep your bowels regular. Staying hydrated and maintaining an active lifestyle can also help with digestion and regularity.
         \item $c_2: $ Oats, beans, lentils, and chickpeas provide soluble fiber. Whole grains, nuts, and vegetables like green beans are good sources of insoluble fiber. Certain fruits like apples and pears also contain both types of fiber.
        \item $c_3: $ Candy is not a good example of soluble fiber or insoluble fiber.
\end{itemize}
In this example clearly $c_1$ answers $q_1$ and $c_2$ answers $q_2$. Yet after calculating embeddings similarities using our text embedder, $d(q_1, q_2) = .05 < d(q_1, c_2) = .16 < d(q_1, c_1) =.17$. This implies that the query $q_1$ is geometrically closer to the semantically similar query $q_2$ than it is to the content $c_1$ that actually contains the answer. A similar issue arises with $d(q_1, q_2) = .05 < d(q_2, c_3) = .14 < d(q_2, c_2) = .15$. This simple toy illustration underscores the limitations of traditional retrieval based solely on embedding similarity.

\section{Content Card Example}
\label{appendix:cc_example}
\begin{mdframed}
\begin{minipage}{0.90\textwidth}
If there is not enough blood on the test strip, you may not get an
accurate blood glucose reading. Some meters won't even give you a
reading at all. So here are some tips to help you get a big enough drop of blood: Rub your hands in warm water to get the blood to your
fingertips. Shake your hand to help force blood to your fingertips.
Hold your hand down by your side for 30 seconds to help blood run to
your fingertips. Set your lancing device to puncture just deep enough
to get the size drop you need. (This may take some trial and error.)
Take blood from the side of your finger. There are fewer nerves there,
so it doesn't hurt as much. If your lancet isn't going deep enough,
dial the number up higher on the lancing device. Once you've lanced,
squeeze your finger where it meets your palm and move toward the tip
of your finger.
\end{minipage}
\end{mdframed}

\section{Analysis of Generated Question Similarity}
\label{gen_q_similarity}
To assess the potential of our question-based retrieval approach, we analyzed the semantic similarity among the generated questions above.
Our goal was to determine if questions derived from the same content card (intra-content similarity) exhibit higher similarity than questions generated from different cards (inter-content similarity).

We calculated pairwise cosine similarity between the generated questions' embeddings. As shown in Table \ref{tab:geom_qc}, mean intra-content similarity (0.871) is indeed higher than mean inter-content similarity (0.827). Similarly, questions are, on average, more similar to their source content card (0.837) than to other content cards (0.762). We note that the relatively high similarity between questions and content from different cards likely stems from the inherent thematic overlap within our dataset, as all content and questions focus on T2D and HTN management.

These results provide initial evidence supporting the premise of our question-based retrieval strategy. While the specific similarity values are influenced by our dataset and embedding model, the consistent trend of higher intra-content similarity suggests that aligning incoming user queries with this question set can improve retrieval accuracy in RAG systems.

\begin{table}[htb]
\centering
\setlength{\tabcolsep}{4pt}  
\scriptsize
\begin{tabular}{ll}
\toprule
\textbf{Average Cosine Similarity b/w} & \textbf{Value} \\
\midrule
Questions generated from the same content & .871 \\
Questions generated from a different content & .827 \\
Questions and their associated content & .837 \\
Questions and other contents & .762 \\
\bottomrule
\end{tabular}
\caption{Cosine similarities between questions and contents combination}
\label{tab:geom_qc}
\end{table}

\section{Benchmark Methods Implementation Details}\label{benchmark-methods}
\begin{enumerate}
    \item \textbf{Naive RAG} \citep{lewis2020retrieval}. 
    This baseline approach retrieves content based on the cosine similarity between the embedded query and candidate content using identical embedders. The query and retrieved contents are then fed into an answer generator LLM.
    
    \item \textbf{HyDE} \citep{gao2022precise}. HyDE uses an instruction tuned LLM that is prompted to generate a hypothetical pseudo-document. This pseudo-document and its embeddings are used for retrieval by searching for similar \textit{real} contents, offloading the relevance modeling to the LLM's generation. Similarly to HyDE, \textbf{Query2Doc} \citep{wang2023query2doc} employs few-shot prompting to generate a hypothetical pseudo-document that would answer the query. Given the similarity with HyDE, we omit Query2Doc from our benchmark.
    
    \item \textbf{QA-RAG} \citep{kim2024rag}. QA-RAG generates a pseudo-answer using a fine tuned LLM. Then, it retrieves one set of content using the query embeddings only, and another set of content using the pseudo-answer's embeddings only. Finally, it uses a Reranker to select only the most relevant contents among this dual set. In our implementation of QA-RAG, we use \texttt{Gemini-Pro} out of the box to generate a pseudo-answer, and retrieve $k$ documents using the pseudo-answer only, without a downstream reranker.
    
\end{enumerate}

\section{Case Study}

To more concretely demonstrate the advantages of QB-RAG in our application, we present an example query from the out-of-distribution test case. The results of Naive RAG, QA-RAG and HyDE for the given test query are as follows.

\smallskip
\noindent\textbf{Test query:}

\begin{mdframed}
\begin{minipage}{0.90\textwidth}
What is the difference between soluble and insoluble fiber and how do they affect the body?
\end{minipage}
\end{mdframed}

\smallskip
\noindent\textbf{Retrieved content from all non QB-RAG methods: }

\begin{mdframed}
\begin{minipage}{0.90\textwidth}
     Fiber is the ``roughage'' in plants that gives them their shape. Your body can’t digest it (it just passes through), so why do we need to eat it? You've probably heard that fiber helps keep you regular. Fiber can also help lower your cholesterol, manage your blood glucose, and keep you feeling full longer. And it can help you lose weight. How much fiber is enough? For women, experts recommend 25 grams of fiber each day. Men need about 35 grams. Ready to add more fiber to your diet? Try a couple of these tips this week: Choose 100\% whole-grain bread or tortillas instead of white bread or white-flour tortillas. Eat brown rice instead of white rice. Fill half your plate with low-carb vegetables. Grab an apple instead of applesauce. Swap in black bean or navy bean soup instead of chicken noodle soup. Choose oatmeal or oat bran instead of corn flakes or puffed rice cereal. Munch on popcorn or kale chips instead of pretzels. Consider trying a fiber-enriched food. A fiber-enriched breakfast bar has around one-third of your daily fiber! A quick note: Fiber is wonderful, but add it to your diet slowly. Your body may need a little time to adjust. As you add fiber, drink more fluids to keep it moving through. If you have any problems with gas or bloating, there are over-the-counter
that can help. Which sources of fiber look good to you?
\end{minipage}
\end{mdframed}

\smallskip
\noindent\textbf{Answer generated using retrieved content from all non QB-RAG methods: }

\begin{mdframed}
\begin{minipage}{0.90\textwidth}
I cannot determine the answer to that. The provided context does not mention the difference between soluble and insoluble fiber or how they affect the body.
\end{minipage}
\end{mdframed}

\smallskip
Given the query about soluble and insoluble fiber, Naive RAG, QA-RAG and HyDE all retrieved a content about fiber as the top content. While on topic, the content does not contain information to differentiate the effect of soluble and insoluble fiber on the human body, leading to the answer generator declining to answer. The results of QB-RAG for the same test query are as follows.

\smallskip
\noindent\textbf{Retrieved query by QB-RAG: }

\begin{mdframed}
\begin{minipage}{0.90\textwidth}
How does soluble fiber help the body? 
\end{minipage}
\end{mdframed}

\smallskip
\noindent\textbf{Mapped content of retrieved query by QB-RAG: }

\begin{mdframed}
\begin{minipage}{0.90\textwidth}
Steel-cut oatmeal (also called Irish oatmeal) is an easy, healthy, high-fiber food. Steel-cut oats have more fiber than rolled or instant oats. Steel-cut oats have both soluble and insoluble fiber. Soluble fiber helps lower cholesterol and keeps blood glucose from rising quickly. Insoluble fiber helps move food through your body, which can keep your bowels regular. The fiber in oatmeal also makes it a very filling food. You can eat a small portion size and feel full longer, which is great for weight loss. Want to boost your oatmeal flavor? Try these ideas: Cinnamon Slivered almonds or other nuts (2 Tbsp) Pumpkin pie spices without sugar Almond butter (1 Tbsp) or other nut butter Powdered peanut butter Powdered protein powder (mix in after cooking for a protein and flavor boost) Stevia (sugar-free sweetener) A dash of vanilla extract Tips: Oatmeal has carbs so keep a small portion size in mind. Milk adds carbs. Consider using a cup of regular milk, unsweetened almond milk, or unsweetened coconut milk as low-carb options. Try eating oatmeal with a protein food (like eggs, nuts, or peanut butter) for a balanced meal.
\end{minipage}
\end{mdframed}

\smallskip
\noindent\textbf{Answer generated using mapped content of retrieved query by QB-RAG: }

\begin{framed}
\begin{minipage}{0.90\textwidth}
Soluble fiber helps lower cholesterol and keeps blood glucose from rising quickly, while insoluble fiber helps move food through your body, which can keep your bowels regular.
\end{minipage}
\end{framed}
\smallskip
Given the same test query, QB-RAG first retrieved a pre-generated query about soluble fiber stored in the knowledge base. It then maps the retrieved query to the content that was used to generate the query. Given the content, the answer generator was able to provide a response to the test query. Note that while the retrieved content is largely about steel-cut oatmeal, it contains information about soluble and insoluble fiber and their effect on the body. On the surface, it may seem that the content retrieved by the other methods is more on topic, QB-RAG was able to pinpoint the exact content that contains the needed specific information with the help of the pre-generated query. 

\section{Prompts}
\label{appendix:prompts}

\subsection{Prompt for QB-RAG Question Generation}

\begin{lstlisting}[numbers=none, caption=Base prompt for question generation, label={lst:base-prompt}]
***SYSTEM:***
You are a Teacher / Professor. Your task is to setup {num_questions} questions for an upcoming quiz/examination. The questions should be both diverse and exhaustive in nature across the document. Restrict the questions to the context information provided.

***INSTRUCTIONS:***
You are presented with a text authored by healthcare professionals, offering advice and strategies for managing conditions such as diabetes and high blood pressure. Your task is to formulate relevant questions that the text is written to address. Closely follow the example questions for style and structure when formulating your own question for the provided text. Your generated questions should be in first person with details, but only at a high school reading level. Your questions should be answerable from the text, but do not copy the text verbatim. MAKE SURE to generate at least {num_questions} questions. Format the generated questions separated by comma in the following JSON format with "questions" as its key: {{"questions": ["...","...","...","..."]}}

***EXAMPLE OUTPUTS:***
Sample Text: {example content}
Generated Questions: {example JSON with list of num_question questions}

--------------------------------------------
Given the context information and no prior knowledge, generate the relevant questions.
Text:
{cc_text}
Generated Questions:
\end{lstlisting}

\subsection{Prompt for Answerability Auto-Evaluator}
\label{prompt:answerability}

\begin{lstlisting}[numbers=none, caption=Prompt for Answerability Auto-Evaluator, label={lst:answerability-prompt}]
***SYSTEM***
You are a health coach providing support for members living with diabetes. You have some basic healthcare and nutrition knowledge.

***INSTRUCTIONS***
Given a pair of user query and a paragraph of content, determine if the content contains relevant information to infer an answer to the query. Think step by step. First provide an explanation, then generate a "Yes" or "No" label. Put the results in a Python dictionary format with keys "Explanation" and "Source relevant".

***EXAMPLES***
For example, given the following query and content as inputs: {positive and negative examples, with explanations}

--------------------------------------------
Provide the output for the following query and content:
Question: {question}
Content: {content}
\end{lstlisting}

\subsection{Prompt for Answer Generation}\label{prompt:answer}
\begin{lstlisting}[numbers=none, caption=Base prompt for answer generation, label={lst:answer-generation-prompt}]
Use only the provided pieces of context to answer the  question at the end. Think step-by-step and then answer. Respond in 3 to 6 short sentences.

Do not try to make up an answer:
- If the context does not contain enough relevant information to determine an answer to the query, say "I cannot determine the answer to that."
- If the context is empty, just say "I do not know the answer to that."

Answer in an empathetic and positive tone. Do not use phrases such as "According to the context", and directly answer the question.

Contexts: {contexts}
Question: {question}
Answer:
\end{lstlisting}

\end{document}